# ENHANCED SKIN COLOUR CLASSIFIER USING RGB RATIO MODEL

Ghazali Osman[1], Muhammad Suzuri Hitam[2] and Mohd Nasir Ismail[3]

[1,3]Faculty of Information Management
Universiti Teknologi MARA, Kelantan, Malaysia
[1]ghaza936@kelantan.uitm.edu.my
[3]nasir733@kelantan.uitm.edu.my
[2]Department of Computer Science
Universiti Malaysia Terengganu, Malaysia
suzuri@umt.edu.my

## ABSTRACT

*Skin colour detection is frequently been used for searching people, face detection, pornographic filtering and hand tracking. The presence of skin or non-skin in digital image can be determined by manipulating pixels' colour and/or pixels' texture. The main problem in skin colour detection is to represent the skin colour distribution model that is invariant or least sensitive to changes in illumination condition. Another problem comes from the fact that many objects in the real world may possess almost similar skin-tone colour such as wood, leather, skin-coloured clothing, hair and sand. Moreover, skin colour is different between races and can be different from a person to another, even with people of the same ethnicity. Finally, skin colour will appear a little different when different types of camera are used to capture the object or scene. The objective in this study is to develop a skin colour classifier based on pixel-based using RGB ratio model. The RGB ratio model is a newly proposed method that belongs under the category of an explicitly defined skin region model. This skin classifier was tested with SIdb dataset and two benchmark datasets; UChile and TDSD datasets to measure classifier performance. The performance of skin classifier was measured based on true positive (TF) and false positive (FP) indicator. This newly proposed model was compared with Kovac, Saleh and Swift models. The experimental results showed that the RGB ratio model outperformed all the other models in term of detection rate. The RGB ratio model is able to reduce FP detection that caused by reddish objects colour as well as be able to detect darkened skin and skin covered by shadow.*

## 1. INTRODUCTION

Skin is the largest organ of human body [1]. It is a soft outer covering of human's muscles, bones, ligaments, and internal organs. Skin colour is produced by a combination of melanin, haemoglobin, carotene, and bilirubin. Haemoglobin gives blood a reddish colour or bluish colour while carotene and bilirubin give skin a yellowish appearance. The amount of melanin makes skin appear darker [2]. Due to its vast application in many areas, skin colour detection research is becoming increasingly popular among the computer vision research community. Today, skin colour detection is often used as pre-processing in some applications such as face detection [3-9], pornographic image detection [10-20], hand gesture analysis [21], people detection, content-based information retrieval, to name a few.

The skin colour fills only a small fraction from the whole colour model and thus, any frequent appearance in an image could be a clue to human presence.  A skin colour classifier defines a decision boundary of the skin colour pixels in the selected colour model based on database of skin-coloured pixels.  Skin colour provides computationally effective, robust information against rotations, scaling, and partial occlusions [22].  Skin colour can also be used as complimentary information to other features such as shape, texture, and geometry.

A simple technique for skin detection modelling is to implement one or several thresholds [23-25] to decide whether a pixel is skin or non-skin.  A more advance modelling technique employing statistical based approaches such as neural network [26], Bayesian [27], maximum entropy [28] and $k$-means clustering [29] have also been used to detect skin colour pixel.  Many different modelling techniques for discriminating between skin and non-skin regions are available in the literature.  The skin distribution modelling techniques can be grouped into four types [30], i.e. explicitly defined skin region, parametric, non-parametric, and dynamic skin distribution modelling techniques.

An explicitly defined skin colour region modelling is perhaps the simplest method often employed by researchers.  This method used to formulate skin detection classifier, which is defined by the boundaries of the skin region in certain colour coordinates in appropriately chosen colour model.  This method is very popular among the researchers [23, 24, 31-38] because it is easy to implement and do not require a training phase.  However, the main problem of this method is, it is difficult to achieve high accuracy in skin detection.

Detecting skin-coloured pixels, although it seems as a straightforward and easy task, but it has been proven to be quite challenging for many reasons.  This is because the appearance of a skin colour in an image depends on the illumination conditions where the image was captured.  Therefore, a major challenge in skin colour detection is to represent the skin colour distribution model that is invariant or least sensitive to changes in illumination condition.  In addition, the choice of colour model used for skin colour detection modelling could significantly affects the performance of any skin colour distribution methods.  Another challenge comes from the fact that many objects in the real world may have almost similar skin-tone colour such as wood, leather, skin-coloured clothing, hair, sand, etc.  Moreover, skin colour is different between human races and can be different from a person to another, even with people of the same ethnicity.  Finally, skin colour will appear a little different when different types of camera are used to capture the object or scene.

The main problem of skin colour detection is to develop a skin colour detection algorithm or classifier that is robust to the large variations in colour appearance.  Some objects may have almost similar skin-tone colour which easily confused with skin colour.  A skin colour can be vary in appearance base on changes in background colour, illumination, and location of light sources, and other objects within the scene may cast shadows or reflect additional light.

Secondly, there are no specific methods or techniques that have been proposed to robust skin colour detection arise under varying lighting conditions, especially when the illumination colour changes.  This condition may occur in both out-door and in-door environments with mixture of day light and artificial light.

Thirdly, many non-skin colour objects are overlapping with skin colour, and most of pixel-based method proposed in the literature cannot solve this problem.  This problem is difficult to be solved because skin-like materials are those objects that appear to be skin-coloured under a certain illumination condition.

In order to enable skin colour detection to cope with the above-mentioned problems, the objective of this study is to enhance a skin colour detection system by using RGB ratio model. To achieve the aforementioned objective, a new skin images dataset have been developed for training and testing. Besides that, a benchmark skin images from Testing Database for Skin Detection (TDSD) [20] and Skin image dataset from Universidad de Chile (UChile) [39] datasets have been used for benchmarking the performance of skin colour classifier that derived from RGB ratio model.

## 2. METHODOLOGY

In general, the purpose of this study is to enhance a skin colour detection using RGB ratio model that able to effectively and efficiently classify the skin and non-skin pixels. To achieve this objective, this section will be described the methodology which is divided in three main steps as follows:

    i.      Data preparation.
    ii.     Skin colour classifiers modelling, and
    iii.    Testing and evaluation.

Algorithm 1 describes in general the whole process that involved in skin colour detection system.

Algorithm 1: Skin colour classification modelling

| | |
|---|---|
| Input: | Skin image. |
| Output: | Skin colour classifier. |
| *Begin* | |
| | Data preparation. |
| | Data transformation from 3D to 2D format. |
| | Formulate RGB ratio |
| | Plot histogram |
| | Derive skin colour classifier |
| | Measure skin colour classifier performance |
| *End* | |

## 2.1. Data Preparation

There are three steps have been involved in data preparation process; skin images collecting from websites, image segmentation and data transformation.

### 2.1.1. Skin Images Collection

There are few skin data images available for public access such as Compaq dataset [40], Sigal dataset [41], Testing dataset for skin detection (TDSD) [20], and db-skin dataset [39]. Researchers such as Jones and Rehg [40], Brand and Mason [31], Brown et al. [42], Jedynak et al. [28], and Lee and Yoo [32] used the Compaq dataset. This dataset consist of 6,818 annotated images. However, at this time of writing, the Compaq dataset is no longer available for public use [39]. Thus, most researchers [23, 24, 29, 39, 43] are using their own dataset.

The Sigal dataset which was developed by Sigal et al. [41] is not suitable to use for skin colour detection algorithm development. This dataset does properly label skin and non-skin regions when labelling the ground truth frames [39]. The TDSD dataset developed by Zhu et al. [20]

consists of 555 images (24 million skin pixels and 75 million non-skin pixels). This dataset consists of many images that are very unsatisfactorily annotated because its annotation process used a semi-automatic process for finding the skin and non-skin ground truth information [39]. Ruiz-Del-Solar and Verschae [39] classified the TDSD dataset images into three groups based on how accurate these images had been marked as skin and non-skin pixels areas, i.e. bad annotated images, good annotated images, and very good annotated images groups. Finally, the dataset called db-skin dataset from Universidal de Chile which consists of 93 still skin images and could be freely obtained via Internet and video images [39]. These images were fully annotated by a human operator. They considered that these images are very difficult to manually segment between the skin and non-skin region because these images have either changing lighting conditions or complex backgrounds containing surfaces or objects with skin-like colours.

Based on limitation of the existing skin colour datasets as mentioned above such as:

i.      Compaq dataset is not longer available, Sigal dataset is not suitable for skin colour distribution modelling,

ii.     Only 100 skin images from TDSD dataset can be used for skin colour distribution modelling because most of skin images are unsatisfied annotation,

iii.    Only 93 skin images can be obtained from UChile dataset, and

iv.     The number of skin images mentioned in (ii) and (iii) are not enough to provide variation in skin colour tone and background colour variation.

Thus, there is a need to develop a new set of skin colour images with reasonable number of skin colour images that has variety of skin tone and background colour. In this study, a new skin colour image database was developed called SIdb (Skin images database) consists of 357 skin colour images which is collected from Corbis website [44] at the royalty free image section. The Corbis website provides a rich resource of skin and non-skin images suitable for content-based information retrieval. It should be noted that images from this website were also used as part of skin images collection by Jones and James [40]. These skin images were divided into two parts, namely a development dataset, which consists of 250 skin images and a testing dataset, which consists of 107 skin images. In other words, the ratio between the development set image and test set image is 70:30.

Throughout in this study, four skin images datasets were used, namely:

i.      Training dataset that consists of 250 skin images. This dataset will be used to develop skin colour distribution model.

ii.     Testing dataset that consists of 107 skin images. This dataset will be used to measure skin colour classifier performance.

iii.    Benchmark image datasets. These datasets also will be used as comparison to measure the performance of skin colour classifier:

    a.      A TDSD dataset that consists of 100 very good annotated skin images [20, 39].

    b.      A UChile dataset that consists 93 skin images of db-skin dataset [39].

## 2.1.2. Image Segmentation

An accurate skin segmentation analysis is considered important in order to have images with the exact ground truth information and to get optimum result in skin detection experiment [39]. It is difficult to obtain good annotated skin and non-skin pixels by using automatic or semi-automatic annotation. Hence, a fully human annotation has been employed.

Each of the images from SIdb was segmented manually using Adobe Photoshop software. This software is widely used for semiautomatic segmentation for anatomical structures in the Magnetic Resonance Images (MRI), Computerised Tomography, other medical images, and for skin and non-skin images segmentation [20].

The Adobe Photoshop software is used under full human annotation. The regions of skin pixels were selected using the Magic Wand tool, which is available in Adobe Photoshop software. This tool enable user to select a consistently coloured area without having to trace its outline. This tool also allows user to interactively segment regions of skin by clicking the area needed. If contiguous area is selected, all adjacent pixels within the tolerance range within the colour region will be selected. The tolerance range defines on how similar in colour of a pixel within the region must be filled. Its value can be adjusted accordingly based on skin image, while regions of skin with complex shape can be segmented quickly. If the region of skin and non-skin are too difficult to segment because of almost skin and non-skin pixels are similar colour, then manual segmentation of skin and non-skin area using pen tracing tool is employed. By using this tool, the user needs to trace skin and non-skin area, manually. Figure 1 illustrates the skin and non-skin annotation to obtain ground truth skin and non-skin information. Meanwhile, Algorithm 2 describes the detail of skin and non-skin segmentation process.

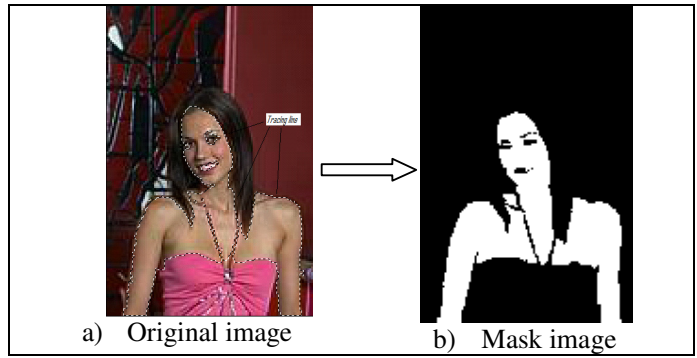

a) Original image  b) Mask image

Figure 1: An annotation process for skin and non-skin
ground truth information using Adobe Photoshop

Algorithm 2: Skin and Non-skin Segmentation

| | |
|---|---|
| Input: | Skin images |
| Output: | Mask images |
| *Begin* | |
| | For all images: |
| | Manually segment skin and non-skin area using Adobe Photoshop software. |
| | Assign skin area pixels value to [255 255 255] |
| | Assign skin area pixels value to [0 0 0] |
| | Save image (Mask image) in Portable Network Graphic (PBG) format |
| *End* | |

This process has to be done carefully to exclude the eyes, hair, mouth opening, eyebrows, moustache and other materials covered on skin area. The RGB value of skin and non-skin areas were mapped to [255 255 255] and [0 0 0], respectively. This process produced a mask image (Figure 1(b)) and stored in Portable Network Graphics (*PNG)* format along with each original image that identifies its skin and non-skin pixels area.

### 2.1.3. Data Transformation

Before skin and non-skin pixels were used for experiments, each pixel of skin and non-skin portion were transformed into 2-dimensional matrix as illustrated in Figure 2. Meanwhile, Algorithm 3 describes the detail of data transformation process. This process involved collecting skin and non-skin pixels from skin images dataset. The collection pixels were done by matching them at corresponding location between original image and its mask image. Each of RGB value (in 3-dimesional matrix) will be transformed to 2-dimesional matrix and labelled as 1 or 0 to indicate whether the pixel belongs to skin or non-skin pixels, respectively.

Algorithm 3: Data Transformation

| |
|---|
| Input:              Skin colour images. |
| Output: 2D matrix data. |
| *Begin* |
|             For all images. |
|         Read images. |
|         Convert 3D matrix images into 2D matrix (3 columns RGB) |
|             Add 4ᵗʰ column Label into 2D matrix |
|             If (pixel == skin) |
|                     Label = 1 |
|             Else |
|                     Label = 0 |
|             End If |
| *End* |

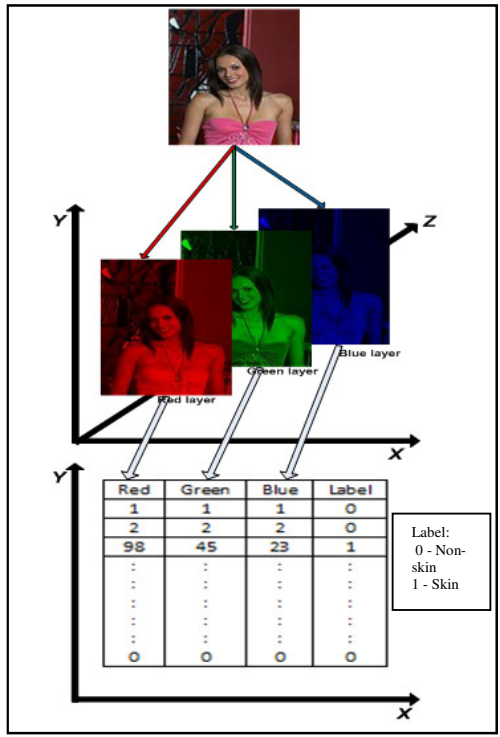

Figure 2: Transformation RGB colour model in
3-dimensional colour layer into 2-dimensional matrix

## 2.2. Skin Colour Modelling

Skin colour distribution modelling is a third step after the choice of colour model has been made and data transformation in skin colour detection algorithm development. In this study, a new technique called RGB ratio model have been introduced. RGB Ratio which proposed in this study is one of the explicitly defined skin region methods. RGB ratio will be formulated by examine and observation from histogram and scatter plot as well as from the Kovac et al. [23], Swift [25], and Saleh [24] models which are based on RGB colour model.

The Kovac model [23] can be divided into four sub-rules as follows:

Pixel is skin colour pixel if:

Rule 1: $R > 95$ and $G > 40$ and $B > 20$ and $\qquad$ (1)
Rule 2: Max $(R, G, B) - \text{Min}(R, G, B) > 15$ and $\qquad$ (2)
Rule 3: $|R - G| > 15$ and $\qquad$ (3)
Rule 4: $R > G$ and $R > B$ $\qquad$ (4)

This rule can be interpreted as the range of $R$ value is from 96 to 255, the range of $G$ value is from 41 to 239, and the range of $B$ value is 21 to 254. Since $R$ value is always greater than $G$ and $B$, the second rule and third rule are always positive values, which can be rewritten as follows:

Rule 2: $R - \min(G, B) > 15$ and $\qquad$ (5)
Rule 3: $R - G > 15$ $\qquad$ (6)

Tomaz et al. [45] described that if $R$-value is too high, and the $G$ and $B$ values are too low, it will result in a pixel is more to red, and should not be considered as skin pixel. In other cases when $R < 100$ and $G < 100$ and $B < 100$, it will result to dark colour that may be non-skin pixel, and when $G > 150$ and $B < 90$ or $R + G > 400$, it will result in yellow like colour. These conditions are not considered in Kovac's rule.

Swift's rule [25] is more simple as compared to Kovac's rule and can be described as follows:

Pixel is not skin colour pixel if:

$B > R, G < B, G > R, B < \frac{1}{4}R$ or $B > 200$ $\qquad$ (7)

The range of $R$-value is from 4 to 255, the range of $B$-value is from 1 to 200, and the range of $G$-value is from 1 to 255. It shows some agreement with Kovac's rule which is that a pixel is considered as skin pixel if $R > G$ and $R > B$ and $B < 200$. However, this rule is still unable to detect some dark skin colour and yellow like colour which is detected as skin colour.

Finally, a very simple rule was introduced by Saleh [24] which consider only the value of $R$ and $G$. This rule defines that a pixel is skin pixel when $R - G$ is greater than 20 and less than 80. That means the range of $R$ is from 21 to 255, while the range of $G$ is from 0 to 234. This rule does not consider a present of $B$-value which contributed to the whitish colour. This rule is also unable to detect dark skin colour or skin cover under shadow, and yellow like colour and redder colour problems which is detected as skin pixel.

By considering the aforementioned issues, a new method has been developed based on painting colour concepts and colour ratio, which is based on colours mixing to produce new colour. This means some ratio of RGB has been taken to develop a new skin colour rule. The sum of $R, G,$ and distance between $R$ and $G$, and $B$ values were observed based on ratio. The histogram of ratio of difference between $R$ and $G$ over the sum of $R$ and $G$, and the ratio of $B$ over sum of $R$ and $G$ are plotted from skin pixel of training dataset as shown in Figure 3. The new rule for skin colour have been developed based on histogram as follows:

$$0.0 \leq \frac{R-G}{R+G} \leq 0.5 \text{ and } \frac{B}{R+G} \leq 0.5 \tag{8}$$

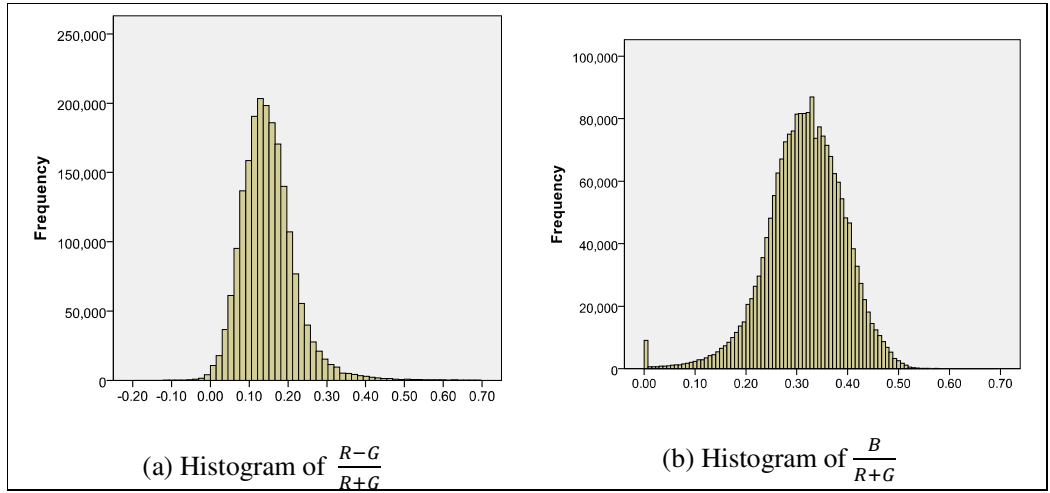

(a) Histogram of $\frac{R-G}{R+G}$        (b) Histogram of $\frac{B}{R+G}$

Figure 3: Histogram of $\frac{R-G}{R+G}$ and $\frac{B}{R+G}$

The performance of classifier formulated from pixel-based classification technique have been compared to the skin distribution model introduced by Kovac et al. [23], Saleh [24], and Swift [25].

## 2.3. Testing and Evaluation

The performance of skin colour detection algorithm can be measured by two methods, i.e. quantitative and qualitative techniques [30]. The quantitative method consists of two techniques, i.e. Receiver Operating Characteristics (ROC) and the true and false positive. Meanwhile, qualitative technique is based on observe the ability of skin colour classifier to classify skin and non-skin pixels from images.

The true positive (TP) and false positive (FP) are statistical measures of the performance of a binary classification test. Binary classification is the task of classifying the members of a given set of objects into two groups on the basis of whether they have some property or not.

The TP is also called sensitivity, measures the proportion of actual positives, which are correctly identified as such. Meanwhile, FP measures the proportion of actual negative which are incorrectly identified. The FP rate is equal to the significance level. The specificity of the test is equal to one minus the FP rate (1 – FP). In case of skin colour detection, the performance of skin colour detection algorithm can be translated to following equation [46]:

$$TP = \frac{I_{pos}}{N_{pos}}, \; FP = \frac{I_{neg}}{N_{neg}} \tag{9}$$

where, $I_{pos}$ is number of skin pixels of testing set correctly detected as skin, $N_{pos}$ is the total number of skin pixels in testing set, $I_{neg}$ is the number of non-skin pixels of the testing set falsely detected as skin, and $N_{neg}$ is the total number of non-skin pixels in testing set. Algorithm 4 describes the skin colour classifier performance measurement using TP and FP.

Algorithm 4:    Skin Colour Classifier Performance Measurement

Input:                 Pixels' features (Records).
Output: True positive rate (TPR) and false positive rate (FPR).
*Begin.*
        N ← No. of skin records.
        M ← No. of non-skin records.
        TP ← 0
        FP ← 0
        For (each record)
                Feed (record into skin colour classifier)
                        If (detected as skin)
                                If (Label ==1)
                                        TP ← TP +1
                                Else
                                        FP ← FP + 1
                                End if
                        End if
        Loop
        TPR ← TP/N
        FPR ← FP/M
*End.*

The performance of skin colour classifiers will be measured using testing dataset, SIdb and benchmark datasets, i.e. UChile and TDSD datasets. The average of performance for these three datasets have been used to indicate overall skin colour classifiers performance. Meanwhile, the qualitative method is measured based on the ability of classifier to detect skin and non-skin on a given skin image.

## 3. RESULT AND DISCUSSION

The explicitly defined skin region method is the easiest and fastest method to define skin colour region. This method is always used as the first step to detect face [23], people and pornographic image [13], etc.

Table 1 shows the performance of the proposed rules as compared to Kovac, Swift, and Saleh rules. Figure 4 illustrates some examples of qualitative result skin colour detection for these rules using qualitative measurement. The white colour indicates as skin pixel while black colour indicates a non-skin pixel.

Table 1: Performance of skin colour classifiers

| Rule | SIdb | | UChile | | TDSD | | **AVERAGE** | |
|---|---|---|---|---|---|---|---|---|
| | TP | FP | TP | FP | TP | FP | **TP** | **FP** |
| Kovac | 90.46 | 10.53 | 81.46 | 16.76 | 93.25 | 24.19 | **88.39** | **17.16** |
| Saleh | 91.50 | 11.66 | 84.40 | 19.08 | 83.46 | 28.12 | **86.47** | **19.62** |
| Swift | 94.16 | 31.27 | 87.67 | 40.24 | 92.83 | 33.63 | **91.55** | **35.05** |
| Proposed | 96.17 | 26.27 | 91.22 | 37.84 | 97.33 | 35.09 | **94.91** | **33.07** |

The RGB ratio method shows the best performance in terms of TP value. It can be concluded that the newly proposed rule can increase the performance of skin colour detection with little big trade-off FP as compared to others.

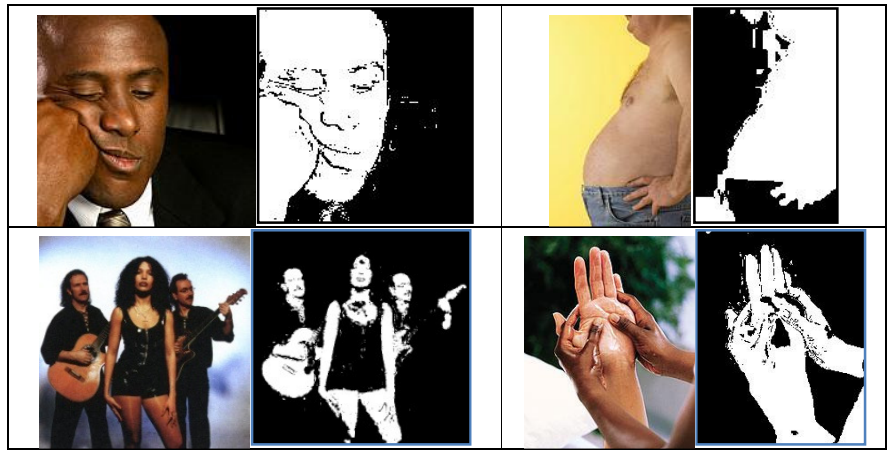

Figure 4: Examples of skin colour classification

## 3. CONCLUSIONS

The explicitly defined skin region method is the easiest and fastest method to define skin colour region. This method is always used as the first step to detect face [23], people and pornographic image [13], etc.

Table 1 shows the performance of the proposed rules as compared to Kovac, Swift, and Saleh rules. Figure 4 illustrates some examples of qualitative result skin colour detection for these rules using qualitative measurement. The white colour indicates as skin pixel while black colour indicates a non-skin pixel.

Table 1: Performance of skin colour classifiers

| Rule | SIdb | | UChile | | TDSD | | **AVERAGE** | |
|---|---|---|---|---|---|---|---|---|
| | TP | FP | TP | FP | TP | FP | **TP** | **FP** |
| Kovac | 90.46 | 10.53 | 81.46 | 16.76 | 93.25 | 24.19 | **88.39** | **17.16** |
| Saleh | 91.50 | 11.66 | 84.40 | 19.08 | 83.46 | 28.12 | **86.47** | **19.62** |
| Swift | 94.16 | 31.27 | 87.67 | 40.24 | 92.83 | 33.63 | **91.55** | **35.05** |
| Proposed | 96.17 | 26.27 | 91.22 | 37.84 | 97.33 | 35.09 | **94.91** | **33.07** |

The RGB ratio method shows the best performance in terms of TP value. It can be concluded that the newly proposed rule can increase the performance of skin colour detection with little big trade-off FP as compared to others.

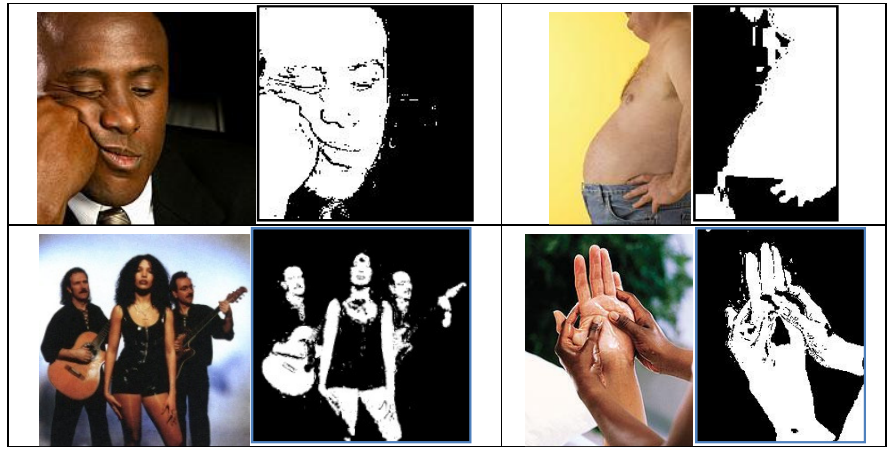

Figure 4: Examples of skin colour classification

## 4. CONCLUSIONS

Skin colour detection is a process to classify a desire pixel into skin or non-skin colour. The classification process can be carried out in two ways, namely pixel-based or region-based classification techniques. The pixel-based skin colour classification sometimes referred to as colour-based skin classification is popular among researchers because it is invariant to scale, occlusion, and rotation. However, one of the problems with pixel-based classification is high false positive (FP), which is a non-skin pixels detected as skin pixel due to similar colour [13].

This study investigated and proposed skin colour distribution model based on pixel-based classification technique using RGB ratio method. The RGB ratio method is categorised as explicitly defined skin region method. This study has successfully achieved the stated objective.

The RGB ratio model has been compared to Kovac, Saleh, and Swift models and the experimental results showed that the RGB ratio model outperform all other techniques in term of TP. Besides using a SIdb dataset that has been developed in this study, the benchmark dataset have been used to test and validate the skin colour classifiers formulated in this study. The benchmark dataset used in this study is TDSD [20] and UChile [39] datasets. The experimental results showed that the classifier formulated also able to detect skin and non-skin from these two benchmark datasets, which produce high true positive (TP) and low false positive (FP). The experimental results also showed that the performances of classifiers are slightly reduced when validate with TDSD dataset which provides slightly high FP. This phenomenon occurred because the TDSD dataset used semi-automatic method to segment skin and non-skin colour, which leads to some disturbances into non-skin colour [39].

The RGB ratio is a new model introduced to explicitly defined skin region model. This model is able to solve some problems related to darken skin colour and skin covered by shadow, which was unable to be detected by other existing skin classifiers. It also can reduce FP rate, which is contributed by reddish objects.

## ACKNOWLEDGEMENTS

We wish to express my sincere appreciation to Professor Javier, Ruiz-del-Solar [39] from Universidad de Chile and Professor Hsu, Rein-Lien [47] from Michigan State University for their skin images database.

## Authors

Ghazali Osman received his B.Sc. from Universiti Kebangsaan Malaysia, his M.Sc. from Universiti Teknologi MARA, Malaysia, and PhD from Universiti Malaysia Terengganu, Malaysia. He is a senior lecture at Universiti Teknologi MARA, Malaysia. His research interests include image processing, artificial intelligent and content-based information retrieval.

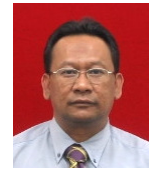

Muhammad Suzuri Hitam is currently a Professor at Universiti Malaysia Terengganu, Malaysia. His main research interests are image processing, artificial intelligence and signal processing.

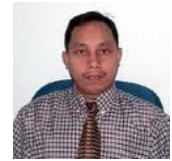

Mohd Nasir Ismail his B.Sc. and M. Sc. from Universiti Teknologi MARA, and PhD from Universiti Sains Malaysia. He is a senior lecture at Universiti Teknologi MARA, Malaysia. His research interests include image processing, artificial intelligent, content-based information retrieval

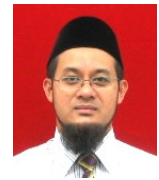